\pdfoutput=1

\documentclass[11pt]{article}
\usepackage{titlesec}
\usepackage[preprint]{acl}

\usepackage{times}
\usepackage{latexsym}
\usepackage{subcaption}
\usepackage{amsmath}
\usepackage{amssymb}

\usepackage{listings}

\usepackage[utf8]{inputenc}
\usepackage{subfiles}
\usepackage{microtype}

\usepackage{inconsolata}

\usepackage{graphicx}
\usepackage{pdflscape}  
\usepackage{array}       
\usepackage{adjustbox}
\usepackage{booktabs}    

\usepackage{multirow}    
\usepackage{ragged2e} 
\usepackage{xcolor}
\usepackage{cellspace} 
\usepackage{enumitem}
\usepackage[font=footnotesize,labelformat=simple]{subcaption}
\usepackage[T1]{fontenc}

\usepackage{hyperref}

\title{Can LLM Agents Maintain a Persona in Discourse?}

\author{
 \textbf{Pranav Bhandari\textsuperscript{1}},
 \textbf{Nicolas Fay\textsuperscript{2}},
 \textbf{Michael Wise\textsuperscript{1}},
 \textbf{Amitava Datta\textsuperscript{1}},
\\
 \textbf{Stephanie Meek\textsuperscript{3}},
 \textbf{Usman Naseem\textsuperscript{4}},
 \textbf{Mehwish Nasim\textsuperscript{1}},
\\
 \textsuperscript{1}School of Physics Mathematics and Computing, University of Western Australia,
 \\
 \textsuperscript{2}School of Psychological Sciences, University of Western Australia,
 \\
 \textsuperscript{3}School of Business and Law, Edith Cowan University,
 \\
 \textsuperscript{4}School of Computing, Macquarie University
\\
 \small{
   \textbf{Correspondence:} \href{mailto:mehwish.nasim@uwa.edu.au}{mehwish.nasim@uwa.edu.au}
 } 
}

\begin{document}
\maketitle
\begin{abstract}



Large Language Models (LLMs) are widely used as conversational agents exploiting their capabilities in various sectors such as education, law, medicine, and more. 
However, LLMs are often subjected to context-shifting behaviour, resulting in a lack of consistent and interpretable personality-aligned interactions. Adherence to psychological traits lacks comprehensive analysis, especially in the case of dyadic (pairwise) conversations. 
We examine this challenge from two viewpoints, initially using two conversation agents to generate a discourse on a certain topic with an assigned personality from the OCEAN framework (Openness, Conscientiousness, Extraversion, Agreeableness, and Neuroticism) as High/Low for each trait. This is followed by using multiple judge agents to infer the original traits assigned to explore prediction consistency, inter-model agreement, and alignment with the assigned personality. Our findings indicate that while LLMs can be guided toward personality-driven dialogue, their ability to maintain personality traits varies significantly depending on the combination of models and discourse settings. These inconsistencies emphasise the challenges in achieving stable and interpretable personality-aligned interactions in LLMs.


\end{abstract}

\section{Introduction}
\begin{figure}[th]
  \centering
  \includegraphics[scale =0.24]{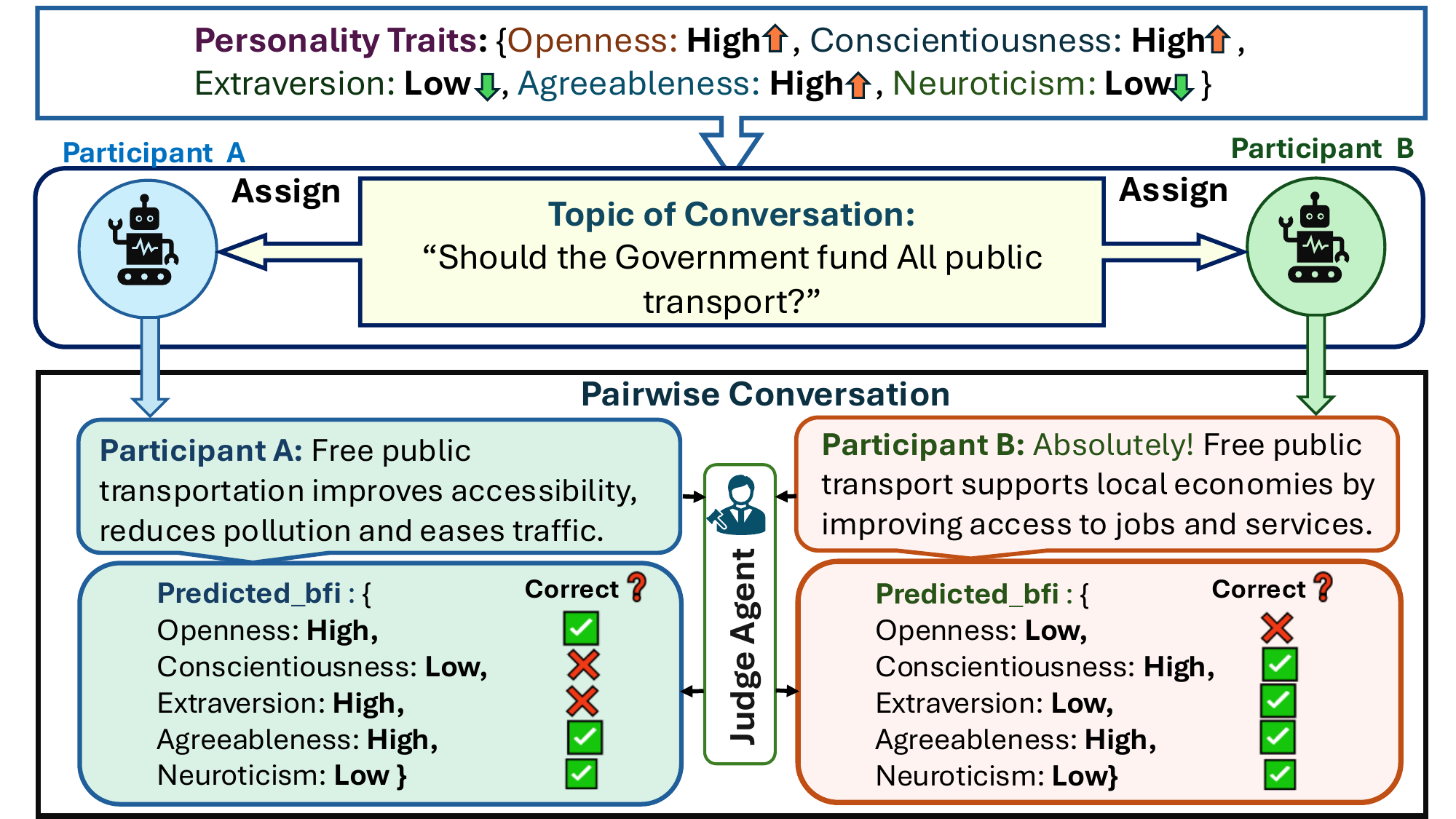}
  \caption{An example of inducing personality in LLM agents, followed by a discourse. A judge agent evaluates whether personality traits were adhered to in the discourse. }
  \label{fig:ToyExample}
  \vspace{-0.35cm}
\end{figure}

Large language models (LLMs) have evolved from task solvers and general-purpose chatbots to sophisticated conversational agents capable of embodying distinct personas. This shift towards personalised agents, driven by LLMs' capacity for perception, planning, generalisation, and learning \cite{xi2025rise}, has enabled context-sensitive discourse and opened up new possibilities across diverse domains.  Persona, defined as conditioning AI models to adopt specific roles and characteristics \cite{li2024steerability}, is a key element in this evolution.  Personalised agents show promise in areas such as emotional support, training, and social skills development \cite{dan2024ptailor}, and are increasingly explored for applications ranging from social science research \cite{zhu2025investigating} to mimicking human behaviour \cite{jiang2023personallm}.  While various personalisation approaches exist, incorporating personas has proven particularly effective in generating contextually appropriate responses and enhancing overall performance \cite{tseng2024twotales, dan2024ptailor}.


Understanding how LLMs express and sustain personality traits in dynamic conversations is crucial, despite their tendency to generate neutral, balanced content. Existing work has explored personality in text using tools like the Big Five Inventory (BFI) \cite{john1991bfi} to infer and analyse personality profiles \cite{bhandari2025evaluating}. However, two key gaps remain. First, it is unclear how consistently LLMs portray assigned personality traits during extended interactions, particularly in pairwise (dyadic) conversations where context shifts and adaptation are necessary. Second, robust methods are needed to evaluate the alignment between the expressed traits in the generated text and the intended psychological profile. We present an example in Figure \ref{fig:ToyExample}. 


While previous studies \cite{jiang2023personallm,kim2025can} have made progress in demonstrating that LLMs can reflect assigned personality traits (often through personality questionnaires), a critical gap remains in understanding how consistently these traits are maintained in generated content, particularly within dynamic conversational settings.  Although assigning personality traits to conversational agents often yields positive results in controlled settings, this does not guarantee that the generated content effectively expresses those traits, nor does it quantify the degree of expression.  Our work addresses this gap by focusing on the generation and evaluation of trait-adherent discourse, specifically within dyadic conversations involving frequent context shifts.  We investigate whether and how LLMs maintain assigned personalities during these dynamic interactions, beyond simply demonstrating the potential for personality reflection to assessing its actual manifestation in conversation.


This work aims to investigate how effectively LLMs express assigned personality traits in generated dialogue.  Specifically, we explore whether and how LLMs maintain Big Five Personality traits, which are represented as the \textbf{OCEAN} framework \cite{husain2025reliability} (\emph{Openness, Conscientiousness, Extraversion, Agreeableness, and Neuroticism}), during dyadic conversations.  We employ a novel agent-based evaluation framework where two LLM agents, each assigned a distinct OCEAN personality profile, engage in a conversation on a given topic.  Subsequently, independent LLM agents (\emph{judges}) assess the generated dialogue to determine the consistency between expressed and assigned traits. This approach allows us to analyse not only whether LLMs reflect personality, but also the peculiarities in trait expression and the challenges of maintaining personality consistency within dynamic conversational contexts.

This work seeks to address the following research questions:


\noindent\textbf{RQ1:} How accurately LLMs as a \emph{judge agent} predict assigned traits from discourse?

\noindent\textbf{RQ2:} How consistently do LLM agents express assigned personality traits in conversations?

\noindent\textbf{RQ3:} Are all OCEAN traits equally prominent in generated conversations?







\begin{figure*}[th]
  \centering
  \includegraphics[scale=0.39]{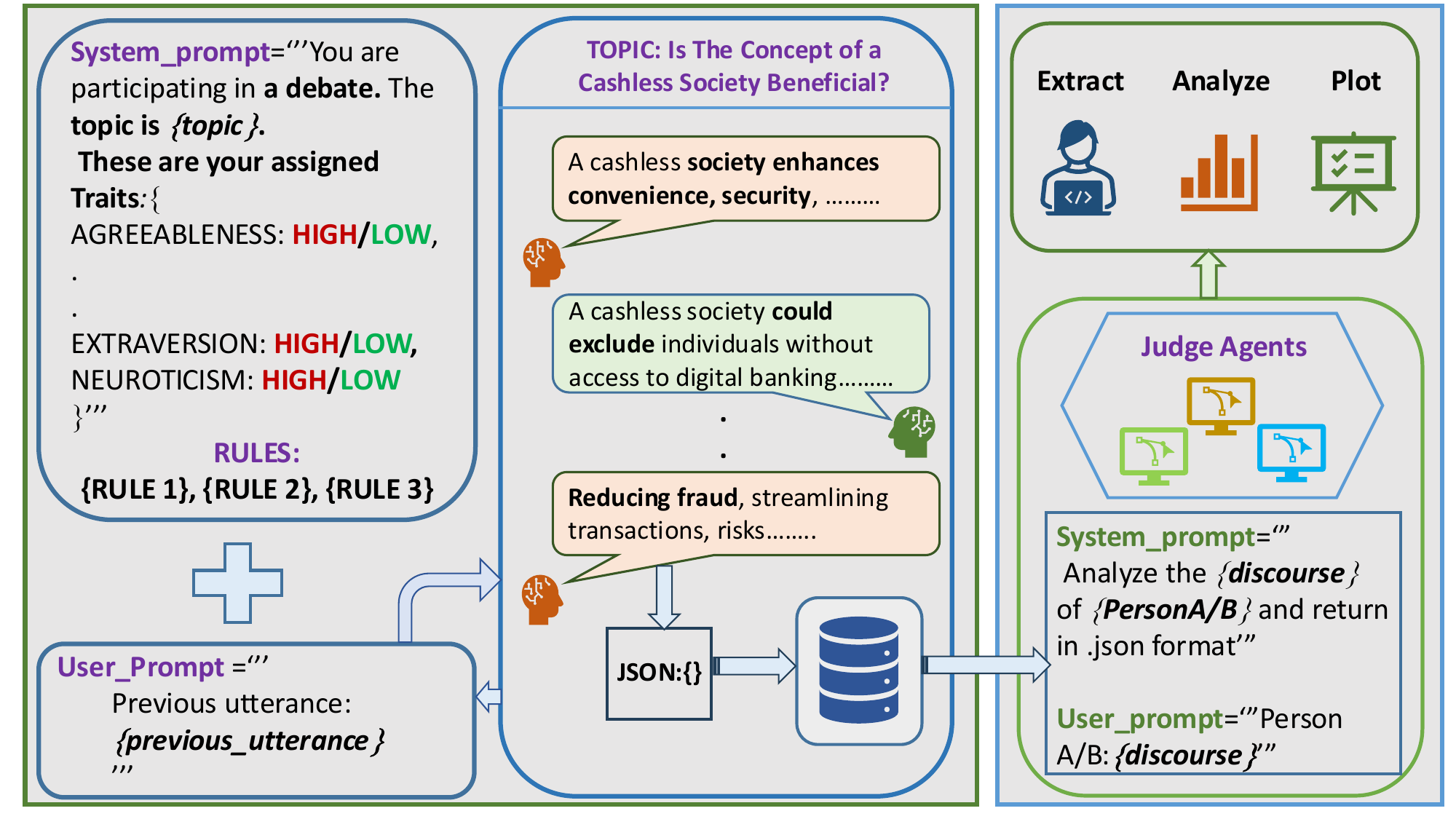} 
  \caption{Methodology of the paper. \textbf{System prompt} inducing traits and topic of discourse are passed with the \textbf{User prompt} containing previous utterance. The conversations are then extracted and analysed by \textbf{Judge Agents} to report the findings.}
  \label{fig:methodology}
\end{figure*}

\section{Related Work}

Personality traits matter since LLMs mimic humans, but their structured psychological evaluation remains an unexplored gap that needs further research \cite{zhu2025investigating}.
The recent literature has looked at designing \cite{klinkert2024evaluating}, improving\cite{huang2024designing}, investigating\cite{frisch-giulianelli-2024-llm,zhu2025investigating}, customizing~\cite{han2024psydial,dan2024ptailor,zhang-etal-2018-personalizing} and exploring \cite{zhu2025investigating,han2024psydial} personality traits. The scope of our work lies both in generating and extracting personality traits embedded within discourse.

\citet{han2024psydial} contribute towards the generation of synthetic dialogues through LLMs. A five-step generation process is used where personality is induced through personality character. Special consideration on prompts is made to infer Pre-trained Language Models (PLM) in generating dialogues. This is because dialogue generation is a challenging task, especially with many constraints and maintaining personality traits. Unlike traditional methods of curating datasets by humans, the authors leverage the capability of PLM to generate synthetic data that is easily scalable. The use of these synthetic datasets significantly improved the ability of LLMs to generate content that is more tailored towards personality traits. While the research is broad, its dataset is limited to Korean and focuses on a single personality trait, which may hinder balanced trait prediction.

While designing and customising the personality traits for LLMs is an intriguing field of study, the focus of this work lies in inducing and investigating the personality traits through discourse generation \cite{yeo-etal-2025-pado}. \citet{jiang2023personallm} investigate the ability of LLMs to express personality traits through essay generation. Using both humans and LLMs as evaluators they explore the personality traits in the generated content. Evaluation through linguistic patterns (LIWC analysis) and human annotation is carried out for GPT models. They show a positive correlation between the generated content and personality traits. However, several gaps are identified such as focusing on closed models, limited data generation and conversations focused on single-ended generation(essays) which does not address the personality expression in scenarios consisting shift of context. Furthermore, the authors suggest models other than OpenAI's GPT models do not follow the instructions well, which results in discarding the content generated by these models for further evaluation. We aim to address this problem through systematic and structural prompting techniques which increases the scope of the analysis.

\citet{sun2024revealing} argue that personality detection should be evidence-based rather than a classification task, enhancing explainability. They introduce the Chain of Personality Evidence (CoPE) dataset for personality recognition in dialogues, addressing state and trait recognition. However, limitations include model specialisation and the availability of a small dataset in Chinese, leaving gaps in the personality trait recognition research.

\noindent\textbf{Prompting methods:} Different methods for assigning personality traits are used in literature, mainly categorising explicit or implicit mention of personality traits or training-based methods. Most studies focus on implementing the OCEAN models to the agents \cite{bhandari2025evaluating, xi2025rise}. One common way of assigning personality traits is through direct allocation of personalities and assigning the personality traits to the agents\cite{}. Another commonly followed methodology is passing content that infers the traits but does not directly mention them \cite{sun2024revealing, han2024psydial}. Personality is also assigned through fine-tuning where distinct fine-tuned models represent distinct personalities. We believe that providing clear instructions about the personas would clear the ambiguity and hence prompt the use of the direct allocation method. 

\noindent\textbf{Evaluation:} LLMs are increasingly used to evaluate personality traits from the text. While their accuracy is still under study, they offer a cost-effective and efficient approach.

\citet{zhu2025investigating} use closed-source models (GPT-4o and GPT-4o-mini) to infer the BFI traits and extract the scores. 

Authors present the findings that the effectiveness of LLMs in predicting personality traits increased as they were prompted with an intermediate step of BFI-10~\cite{bfi10} questionnaires. Two main metrics were used to benchmark the ability of LLMs: correlation and mean difference, where correlation measured the ability to capture structural relationships and mean difference captured absolute prediction accuracy. We also adapt these metrics to evaluate the content produced by LLMs in our agent ecosystem. Different validation datasets relating to personality traits include: Essay Dataset~\cite{yeo-etal-2025-pado}, myPersonality~\cite{mypersonality}, and Twitter Dataset~\cite{twitterdata}.

In summary, the main problems identified in the literature are the use of closed-source models, the lack of analysis in content generation consisting of context-shifting behaviour, and the lack of use of standard evaluation metrics. Furthermore, one of the main challenges in incorporating personality traits is understanding whether all five traits are effectively adhered to in the content that is produced. We aim to address some of these problems through this research.

\section{Methodology}

We present the methodology of this work in Figure \ref{fig:methodology}. In an agent-based setting the methodology is operationalised in 4 phases: 
\emph{\textbf{P}ersonifying agents}, 
\emph{\textbf{G}enerating discourse}, 
\emph{\textbf{E}xtracting personality within discourse}, and 
\emph{\textbf{E}valuation}.
A detailed explanation of the modular approach is presented in subsequent sections.
In summary, the \emph{psychological personas} are assigned to two agents and asked to converse on a topic. The discourse is evaluated using independent agents --- \emph{judge agents} through several evaluation metrics.

We adopted an iterative approach to refine the methodology. Various problems were encountered while producing the discourse between the models, starting with synchronization issues, over-generalisation, repeating the prompts, and explicitly mentioning the personality that the LLMs have assumed. 
Furthermore, in a dyadic conversation between two agents, the subsequent dialogues are highly dependent on the previous conversation, hence one unjustified/bad response can cause the whole conversation to deviate from its original objective. 
Hence, special consideration has been given to achieving complete and sensible conversations.
To validate that LLMs are not generating the same dialogues as before, we perform a similarity check across all the dyadic conversations and validate them.

We selected GPT models from OpenAI\cite{openai2024gpt4omini} and LLaMA models from Meta\cite{llama_cite} due to their popularity and reach. As the landscape rapidly evolved, we expanded our scope to include DeepSeek\footnote{\href{https://huggingface.co/deepseek-ai/deepseek-llm-67b-chat}{DeepSeek models}} to ensure broader coverage and comparison across architectures.
 
Since the generation of essays on a particular topic has been explored in literature such as \cite{kim2025can,yeo-etal-2025-pado}, we wanted to explore the generation of discourses, particularly for two reasons \textbf{1)} The complexity of the topic increases and maintaining a progressive discussion given the explicit persona is a difficult task. \textbf{2)} It is also interesting to understand the consistency in the personality during a conversation. 


\noindent\textbf{Dataset:} We have carefully selected 100 different topics that require, ethical, moral, social or political considerations
\footnote{\href{https://tpd.edu.au/most-controversial-debate-topics/?srsltid=AfmBOooSOI5B5SmGeneZVV9jMuIyqskncYVGpKYepmcgntSW15Czt6Vb}{Debate Topics}} and 20 different combinations of random traits (more in Appendix).

\subsection{Prompt formation} \label{sec:prompt}

There are two basic requirements to create the discourse between two agents. The first one is the assigned persona of the OCEAN model 
(the Big Five Inventory)~\cite{john1991bfi} that is to be maintained at all times while producing an utterance and second is the consideration of the previous utterance in the dyadic conversation so that the current utterance reflects the understanding of the previous utterance and is not an independent reply. In addition, the context of the utterances must be lexically similar to the topic given. 

The prompt formation is an essential part of our methodology. Since the discourse is analysed by other agents and we draw the results based on the discourse, it must be structured robustly to ensure reliability and objective evaluation.

Prompting for LLMs is carried out through specific prompting methods where agents are assigned roles to convey requirements and expected outcomes. Usually, the \emph{system and user} roles are passed as arguments \cite{yeo-etal-2025-pado} in which the system role is responsible for defining the behaviour and limiting the scope of response and the user role is used for defining the input. Despite strict adherence to these techniques, agents may still be overwhelmed by excessive constraints. 

\noindent\textbf{System Prompt}: The system prompt in our case contains the rules for debates carried out on a specific topic. Structured prompts enhance clarity for agents, improve effectiveness, and help users create inclusive prompts despite multiple constraints. Although the formatting of the prompts varies according to the model specifications, they contain the following information. 
\begin{itemize}[noitemsep,leftmargin=*]
    \item The traits are assigned in two forms of extremities: \emph{High or Low}. 
    \item You are a participant in a discourse in which the topic is \emph{${topic}$} and presented with the following traits \emph{${traits}$}.
    \item Assigned personality traits must be maintained throughout the conversation but not explicitly mentioned in the utterances.
    \item Each utterance must be under 50 words and the previous utterance needs to be addressed.
\end{itemize}


\noindent\textbf{User Prompt:} User prompt in this case contributes to an important role in shaping the conversation because the previous discussions are passed through the user prompt to generate the next utterance. 


During the experiments, we noted that GPT models followed instructions effectively in a zero-shot setting with minimal guidance, while models like Llama and DeepSeek required more detailed explanations and constraints. This suggests that GPT models are more adaptable to imperfect prompts compared to other state-of-the-art models.

\subsection{Validation} \label{sec:validation}
Validation involves both human assessment and agent-based evaluation. Discourse quality and coherence are checked via:  \textbf{1)} A human observation of 10-15 discourses is made randomly for each of the categories for the length, content, coherence and quality of the discourse. \textbf{2)} For each course of discourse, we analyse the similarity scores between all the utterances to make sure that the same arguments are not repeated. 
LLMs are used in the literature for personality trait extraction \cite{zhu2025investigating, sun2024revealing}. We employ PLMs to analyse dialogues to infer personality traits and then use pre-assigned personality traits as ground-truth data for evaluation in Section \ref{sec:evaluation}.

\setlength{\cellspacetoplimit}{12pt}
\setlength{\cellspacebottomlimit}{12pt}
\newcolumntype{C}[1]{>{\centering\arraybackslash}m{#1}}
\newcolumntype{L}[1]{>{\RaggedRight\arraybackslash}m{#1}}

\definecolor{headerblue}{RGB}{25,113,194}
\definecolor{rowgray}{RGB}{240,240,240}


\begin{table*}[h!]
    \centering
    \renewcommand{\arraystretch}{1.1}
    \footnotesize  
    \begin{tabular}{@{}L{.2cm}C{0.3\textwidth}C{0.3\textwidth}C{0.3\textwidth}@{}}
        \toprule
        \textbf{\color{headerblue}Judge} & 
        \textbf{\color{headerblue}GPT-4o vs GPT-4o-mini} & 
        \textbf{\color{headerblue}GPT-4o vs LLaMA-3.3-70B-Instruct} & 
        \textbf{\color{headerblue}GPT-4o vs DeepSeek} \\
        \midrule
        
       \rotatebox{90}{GPT-4o} & 
        \includegraphics[width=\linewidth]{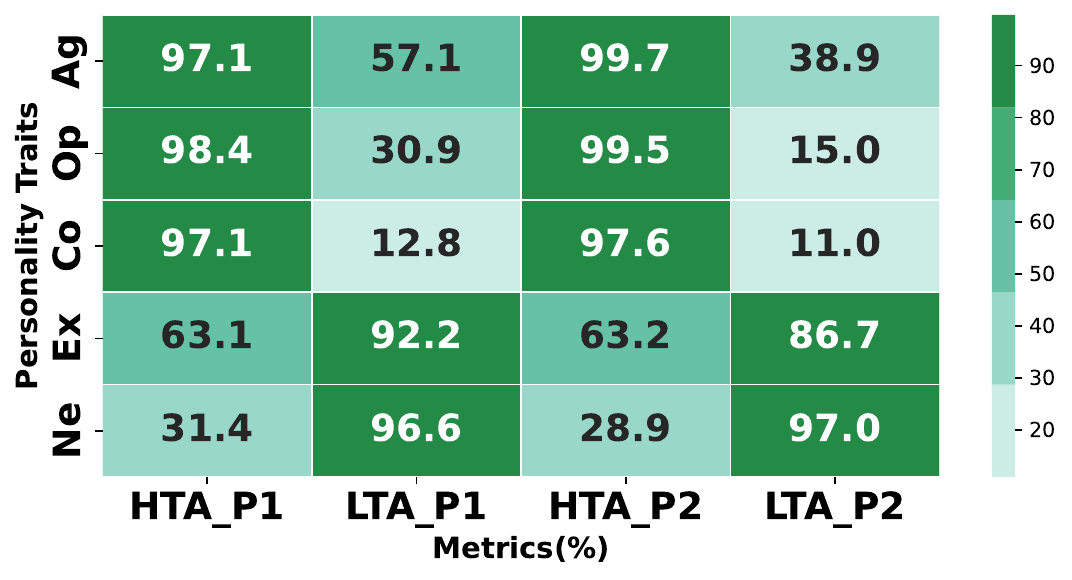} & 
        \includegraphics[width=\linewidth]{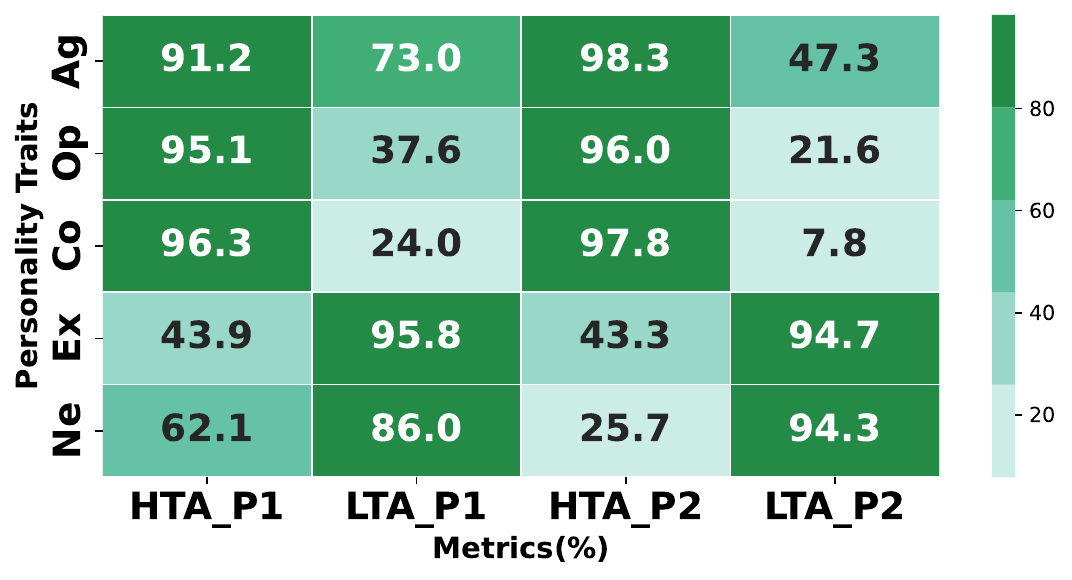} & 
        \includegraphics[width=\linewidth]{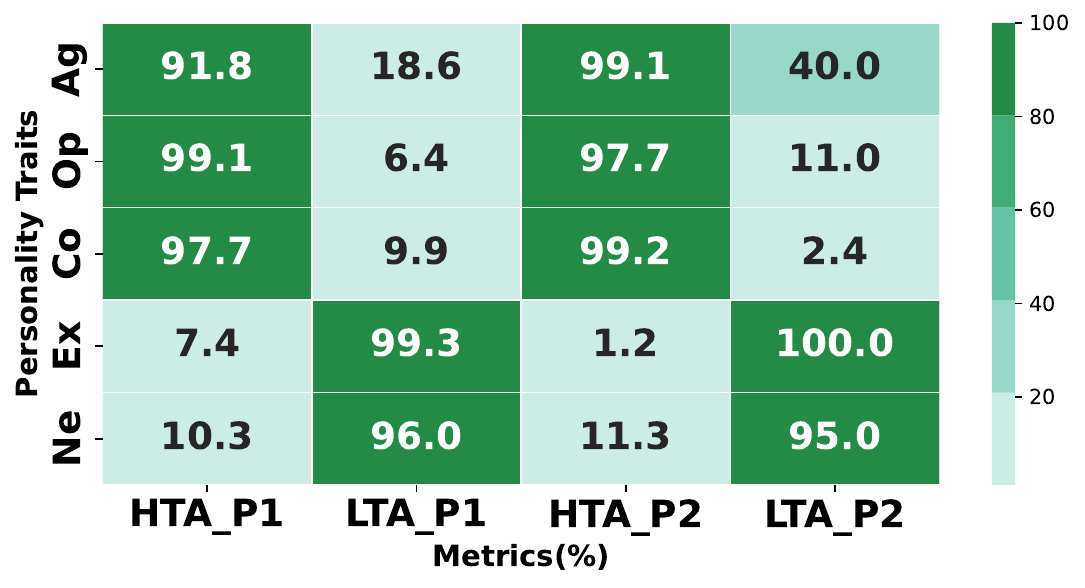} \\
        \addlinespace[3pt]
        
        \rotatebox{90}{GPT-4o-mini} & 
        \includegraphics[width=\linewidth]{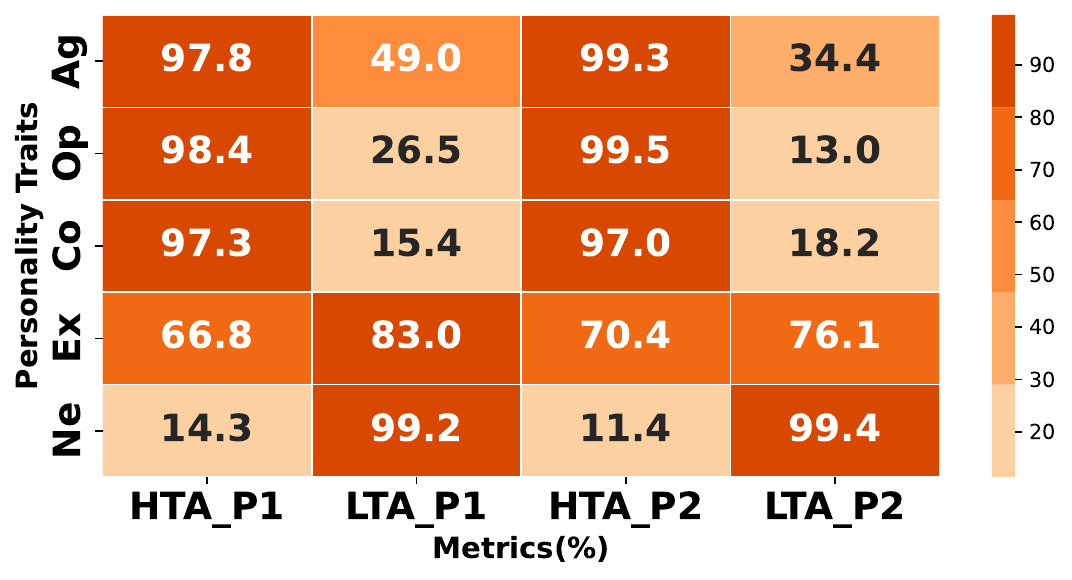} & 
        \includegraphics[width=\linewidth]{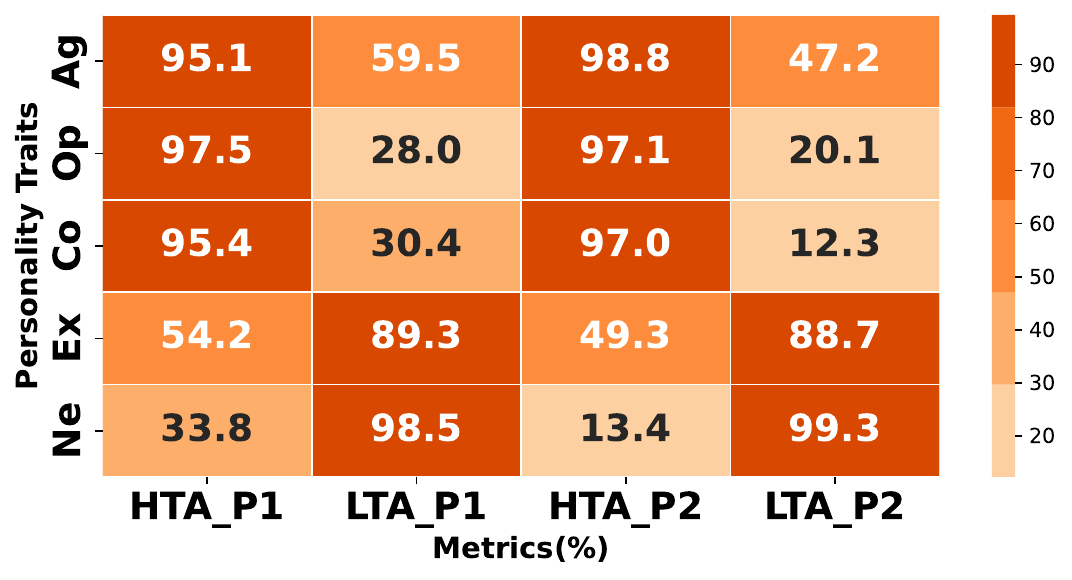} & 
        \includegraphics[width=\linewidth]{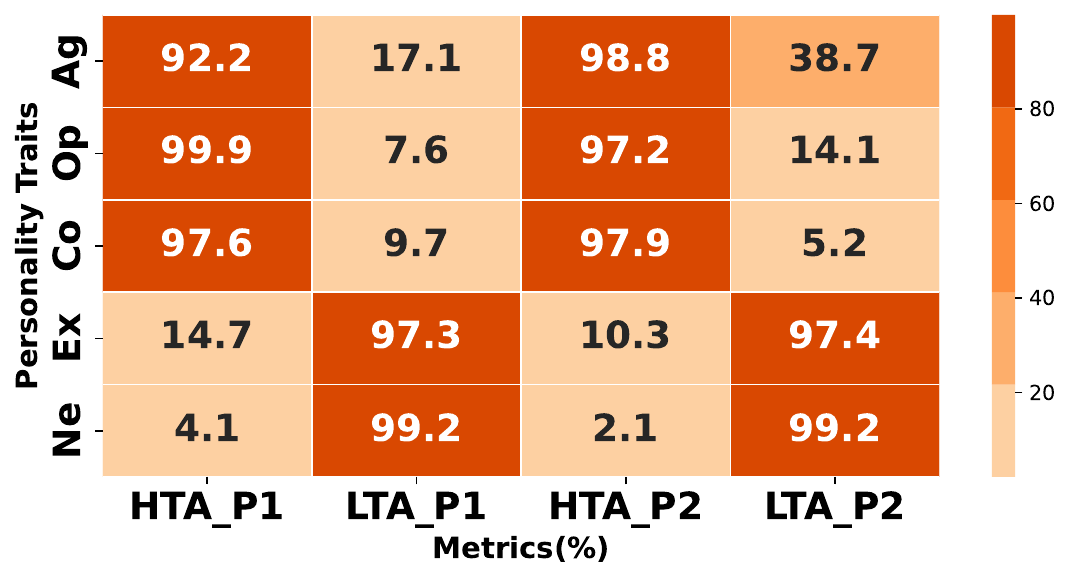} \\
        \addlinespace[3pt]
        
        \rotatebox{90}{LLaMA} & 
        \includegraphics[width=\linewidth]{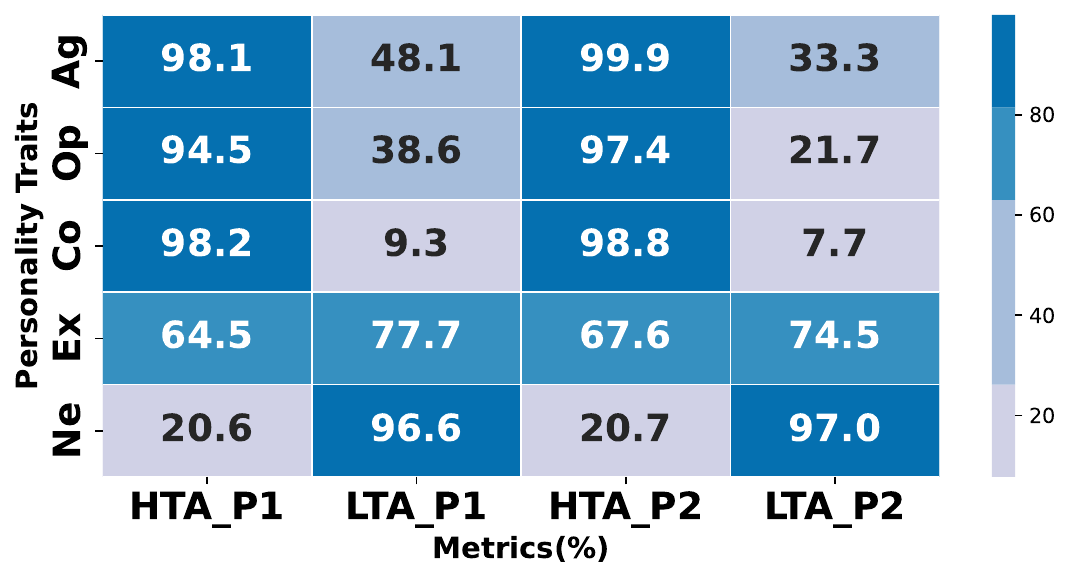} & 
        \includegraphics[width=\linewidth]{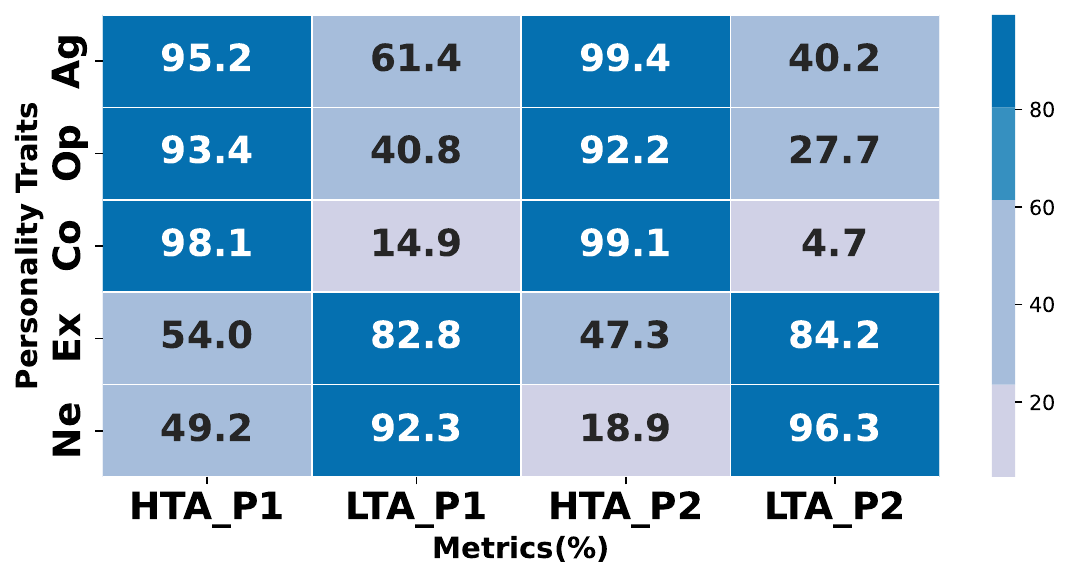} & 
        \includegraphics[width=\linewidth]{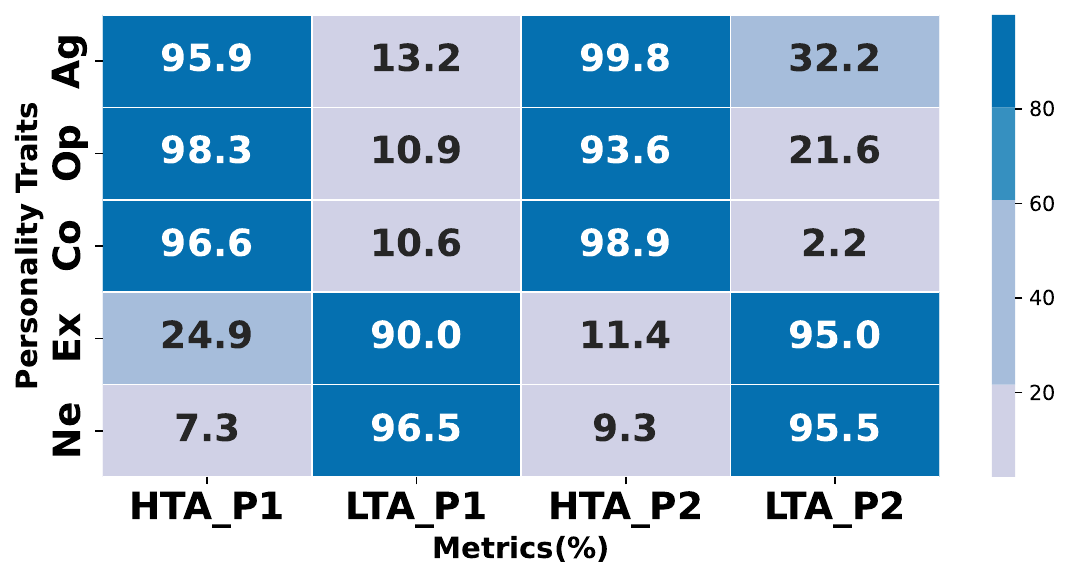} \\
        \addlinespace[3pt]
        
        \rotatebox{90}{Qwen} & 
        \includegraphics[width=\linewidth]{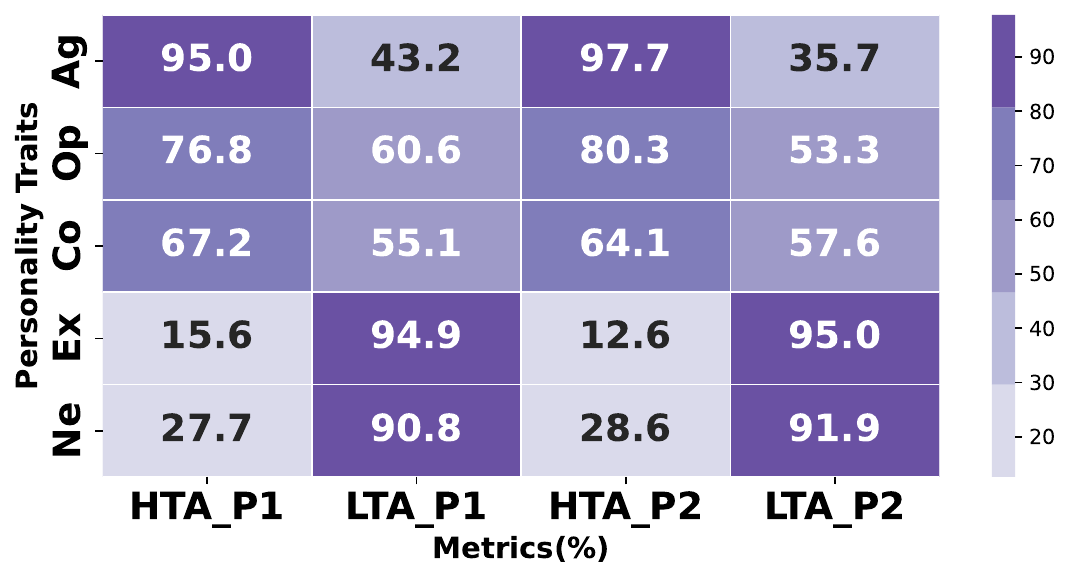} & 
        \includegraphics[width=\linewidth]{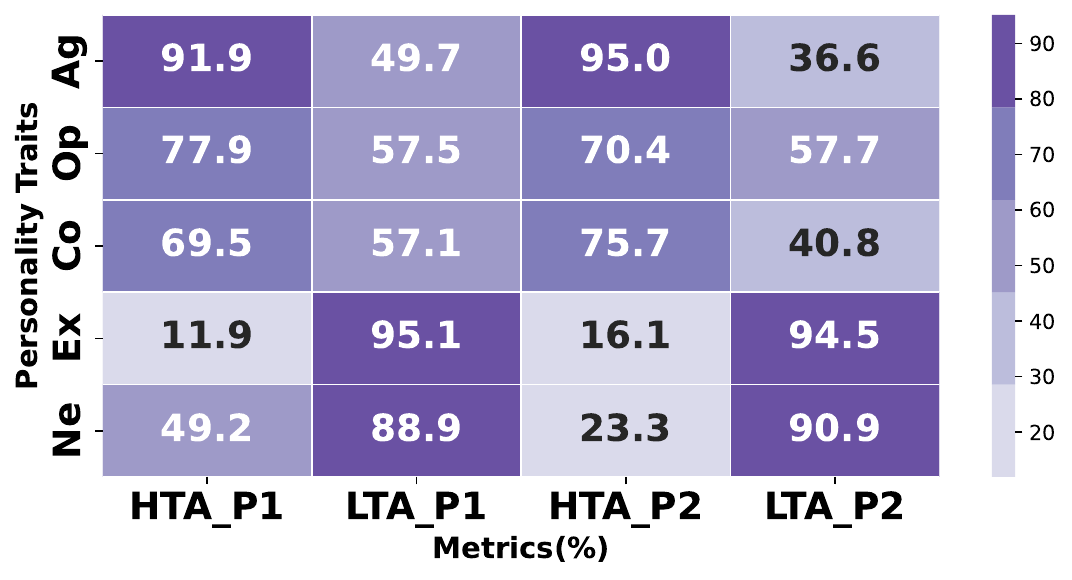} & 
        \includegraphics[width=\linewidth]{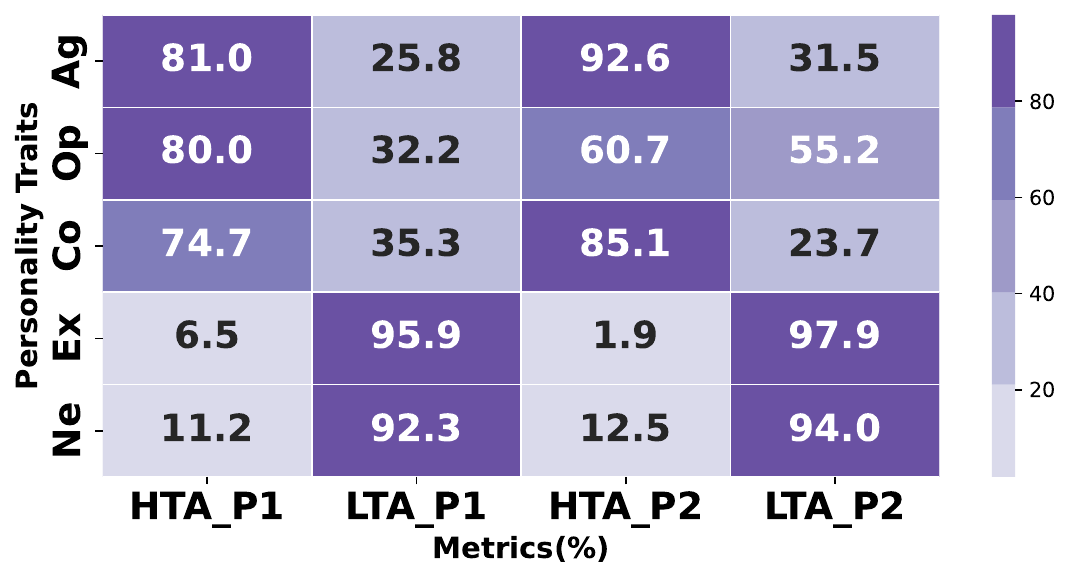} \\
        \bottomrule
    \end{tabular}
    \vspace{-0.2cm}
     \caption{Calculation of High Trait Classification Accuracy\textbf{(HTA)} and Low Trait Classification Accuracy\textbf{(LTA)} for \textbf{Participants 1 and 2} across all the conversations for all the \textbf{Judge Agents}.}
     \label{tab:comparison}
\end{table*} 



\section{Evaluation} \label{sec:evaluation}
Once the discourses are generated, each of the discourses is evaluated by \emph{Judge agents}. The judge agents return data in a \emph{json} format with their prediction of each speaker's personality traits in the text. To reduce the bias of human vs agent-generated content, we provide the utterances to the Judge agents specifying that they are `human-generated'. The following evaluations are made:


    \subsection{Personality prediction consistency Across Models:}

    Personality prediction consistency Across Models:
    With access to both the assigned traits (Section \ref{sec:prompt}) and inferred traits (Section \ref{sec:validation}) using different judge agents, we begin by calculating the accuracy of the models' predictions (a.k.a. inferred traits). We calculate the accuracy of prediction in two different ways: the accuracy of predicted \emph{High} for each trait as High Trait Classification Accuracy(HTA) and finally accuracy of predicted \emph{Low} for each trait as Low Trait Classification Accuracy(LTA). Recall, that we assign a high or a low value for each \emph{OCEAN} trait while assigning personalities in Section \ref{sec:prompt}. We create a confusion matrix for this labelling all the True and False predictions of High and Low values to compute the HTA and LTA values. 

    HTA measures how well the models classify traits assigned as High originally. This is computed by creating a confusion matrix for correct and incorrect classifications. HTA is calculated by dividing the total correctly classified High by the total number of High cases. 

    LTA on the other hand measures how well the models classify traits assigned as Low originally. It is calculated by dividing the total correctly classified Low by the total number of Low cases. An important aspect of this study is understanding potential bias in classification into High or Low traits. While overall accuracy may be high, we focus on whether both categories are proportionately represented.

    \subsection{Inter-rater reliability among the models:} Inter-rater reliability is the measure to understand the agreement between the models. Kappa statistics($\kappa$) is a common method to assess the consistency of ratings among raters (Judge LLMs) \cite{perez2020systematic}. 
    
    We computed Fleiss' Kappa by first gathering personality trait predictions from five different judge models. Each model analysed debates across multiple topics and rated Big Five personality traits for two participants (P1 \& P2). We structured the data so that all model ratings for the same Topic-Trait pair were aligned, ensuring consistency in comparison. After validation, we reformatted the dataset into a matrix where each row represented a topic-trait combination. The matrix contained counts of how many models classified the trait as \emph{High} or \emph{Low} for both P1 and P2 separately. We calculated the inter-model agreement for each trait using Python’s `statsmodels'\footnote{\href{https://www.statsmodels.org/stable/index.html}{statsmodels}} package, specifically the fleiss\_kappa function to extract the consistency of various judge models across all topics.  

    While the first measure explores the accuracy with which the models correctly identify \emph{High} and \emph{Low}, respective to the ground values, this method explores the agreement between the models for a particular trait at a time, irrespective of the base values. 

    \subsection{Discourse alignment with Assigned Personality Traits:} The discourse alignment with assigned personality traits is an important part of this analysis as it depicts if the personality traits are reflected in the contents generated by the agents. We analyse if the discourses linguistically align with the assigned personality traits. Various factors like language, tone and argument structures contribute towards the alignment of personality traits with the content produced \cite{pennebaker1999linguistic}. 
    Linguistic Inquiry and Word Count (LIWC-22)\cite{boyd2022development} analysis is a widely used tool for this category that classifies words into psychological and linguistic categories. \cite{ireland2014natural} explain how natural language and linguistic markers can effectively serve as an indicator of personality traits. For instance, extroverts tend to use more positive words and social process words to reflect their sociable nature. 
    Linguistic markers are successfully able to understand and predict the personality traits in given text \cite{mairesse2007usinglinguistcmarker}. We use the capabilities of LIWC-22 to extract the linguistic features and systematically map the five personality traits from the data to analyse the results.

\section{Results}

The experiments are carried out in two phases: \textbf{1).}
Agents are personified and discourse is generated on a given topic; \textbf{2).} Personality traits are extracted from the discourses and evaluation is performed. This evaluation is critical for determining the controllability of personality traits in language models and validating their alignment with intended psychological characteristics.


Four models are involved in the creation of discourse in different combinations (GPT-4o vs. GPT-4o-mini, GPT-4o vs. Llama-3.3-70B-Instruct, GPT-4o vs. Deepseek-llm-67B-Chat). All of these models have been set up at higher temperatures ($>$0.8) to allow creativity during discourse generation. Limited by resources(NVIDIA A6000 GPU), the larger models such as  Llama-3.3-70B-Instruct and Deepseek-llm-67B-Chat, were quantized to generate discourse. The max\_tokens were limited to 150 to prevent the model from generating verbose utterances.

For the evaluations of the generated discourse, we used {five} different models: GPT-4o, GPT-4o-mini, Llama-3.3-70B-Instruct, Qwen-2.5-14B-Instruct-1M, and Deepseek-llm-67B-Chat --- \emph{the judge agents}. The idea is to include a variety of models(both small and large) and understand the consistency in the results.  

Utterances from LLaMA-3.3-70B-Instruct and DeepSeek-LLM-67B-Chat required filtration due to prompt repetition and inline tags whereas GPT models adhered to instructions effectively. 
\begin{table}[h]
    \centering
    \small 
    \setlength{\tabcolsep}{4pt} 
    \resizebox{0.5\textwidth}{!}{ 
    \begin{tabular}{lcccccc}
        \toprule
        \multirow{2}{*}{\textbf{Trait}} & \multicolumn{2}{c}{\textbf{Discourse 1}} & \multicolumn{2}{c}{\textbf{Discourse 2}} & \multicolumn{2}{c}{\textbf{Discourse 3}} \\
        \cmidrule(lr){2-3} \cmidrule(lr){4-5} \cmidrule(lr){6-7}
        & \textbf{P1} & \textbf{P2} & \textbf{P1} & \textbf{P2} & \textbf{P1} & \textbf{P2} \\
        \midrule
        Agr   & 0.500 & 0.557 & 0.242 & 0.692 & 0.518 & 0.532 \\
        Ope   & 0.699 & 0.420 & 0.534 & 0.631 & 0.250 & 0.430 \\
        Con   & 0.352 & 0.366 & 0.502 & 0.421 & 0.330 & 0.367 \\
        Ext   & 0.123 & 0.097 & 0.235 & 0.105 & 0.287 & 0.260 \\
        Neu   & 0.480 & 0.293 & 0.233 & 0.463 & 0.351 & 0.389 \\
        \bottomrule
    \end{tabular}
    }
        \caption{Fleiss' Kappa Scores for Personality Trait Agreement. \emph{Discourse 1 :} \textbf{GPT-4o vs. GPT-4o-mini}, \emph{Discourse 2}: \textbf{GPT-4o vs. Llama-3.3-70B-Instruct} and \emph{Discourse 3}: \textbf{GPT-4o vs. Deepseek-llm-67b-chat}. P1 and P2: Participants 1 and 2 respectively. }
    \label{tab:fleiss_kappa}
\end{table}
    
\subsection{Personality Prediction Consistency across models} Figures in Table \ref{tab:comparison} represent the result of personality prediction for each of the Judge models. We now describe various interesting patterns observed with different models as Judges. 
   
   \textbf{Analysis across judge models}:
    We note that for Agreeableness, Openness, and Conscientiousness, GPT-4o, GPT-4o-mini, and LLaMA-3.3-70B-Instruct achieve comparable and high-quality results for both Person 1 and Person 2, exceeding 90\% accuracy. However, for the same categories of traits, Qwen-2.5-14B-1M produces significantly low numbers for Openness and Conscientiousness while the scores for Agreeableness are comparable. From the perspective of the size of the models, larger models (GPT-4o and Llama-3.3-70B-Instruct) have higher accuracy in predicting the \emph{High} classification compared to smaller models (Qwen-2.5-14B-Instruct-1M). However, the accuracy of predicting the \emph{Low} trait was significantly high for Openness and Conscientious with Qwen-2.5 as a Judge as compared to other models for both persons 1 and 2. Overall, for Agreeableness, Openness and Conscientiousness the ability of the (GPT-4o, GPT-4o-mini and Llama-3.3)models to predict their High values is significantly higher than predicting the Low values.

\begin{figure*}[t]
    \centering
    \subfloat[GPT-4o vs. GPT-4o-mini\label{fig:first}]{        \includegraphics[width=0.3\textwidth]{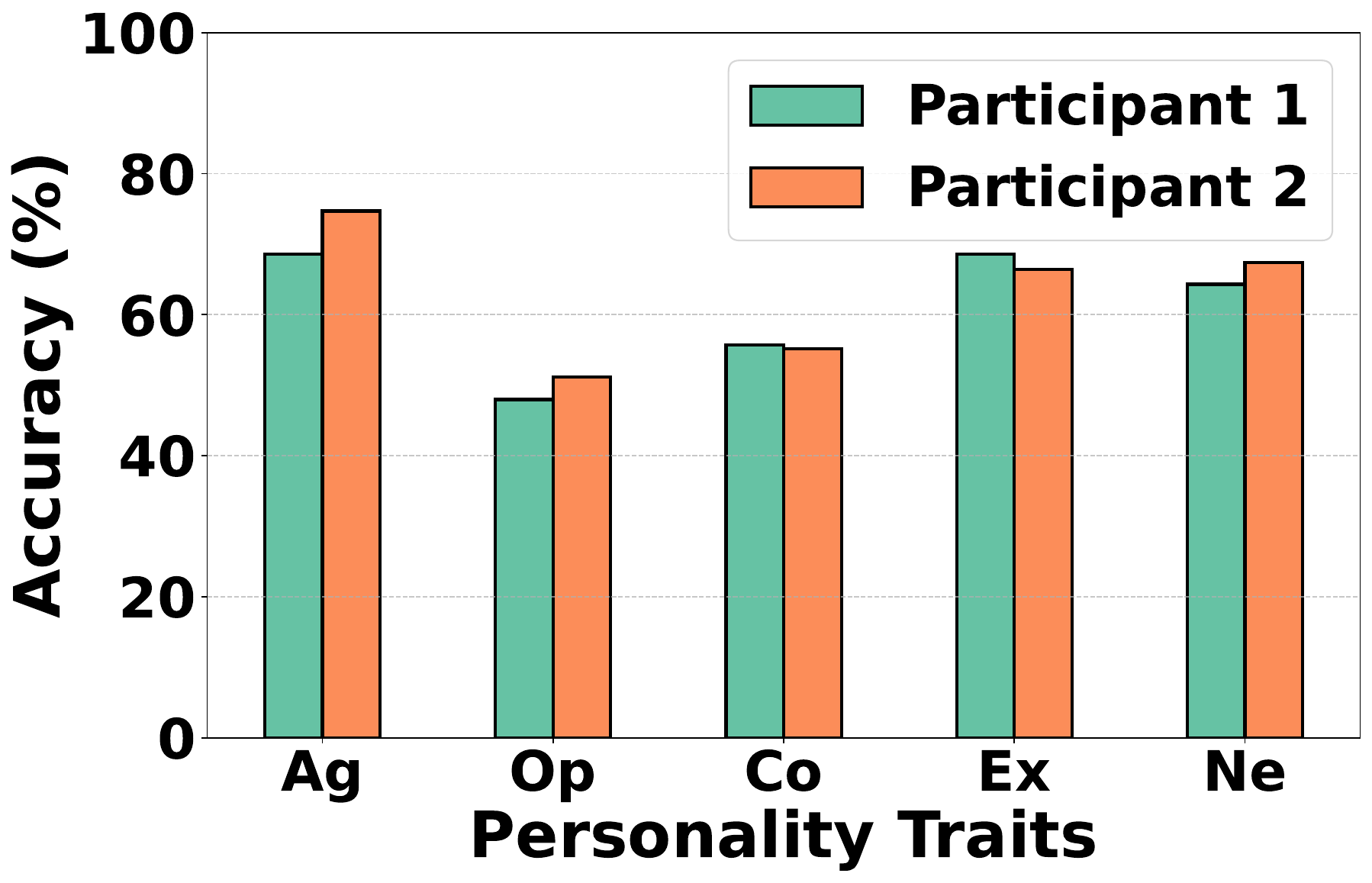}}
    \hfill 
    \subfloat[GPT-4o vs. LLaMA-3.3\label{fig:second}]{
        \includegraphics[width=0.3\textwidth]{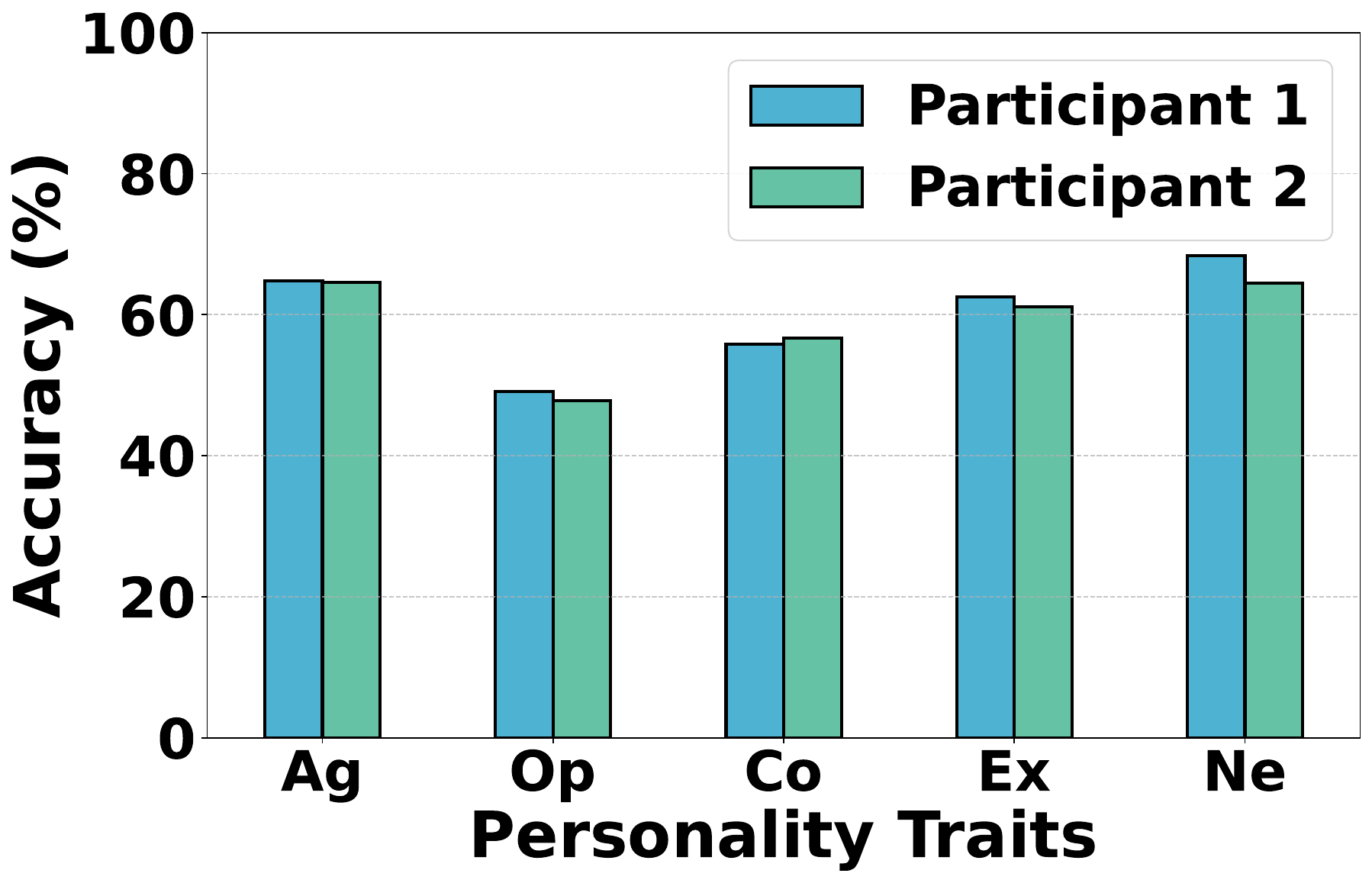} }
    \hfill 
    \subfloat[GPT-4o vs. DeepSeek\label{fig:third}]{
        \includegraphics[width=0.3\textwidth]{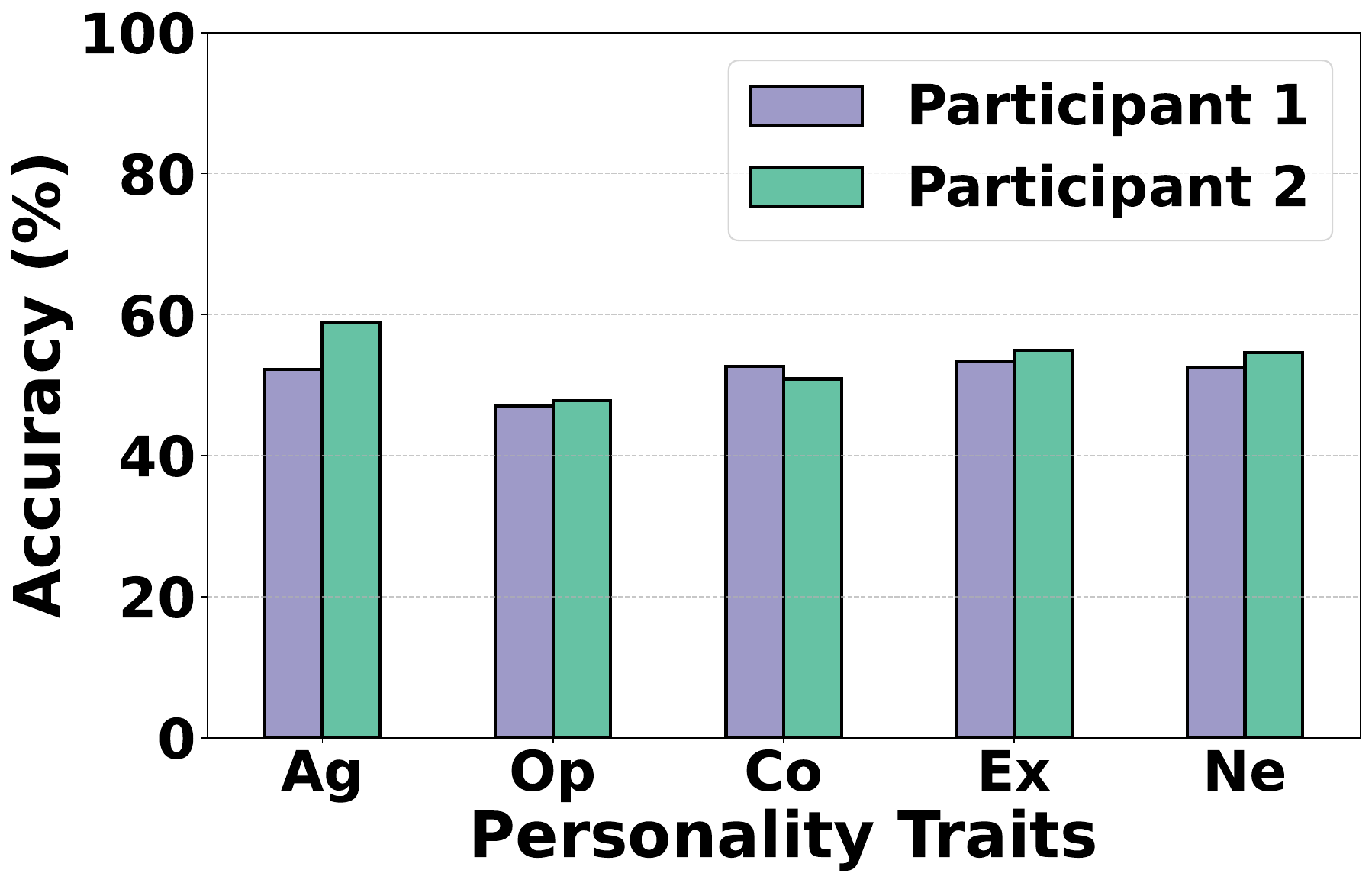}}
    \caption{LIWC analysis depicting the accuracy of conveying the assigned personality traits to Participants 1 and 2.}
    \label{fig:three_figures}
\end{figure*}

        Judgments for Neuroticism and Extraversion show a distinct pattern, with High values predicted less frequently across all discourses and participants. When observed with scrutiny, detecting High Neuroticism is particularly challenging, likely due to judge models failing to recognise it in text or conversational models avoiding highly neurotic responses. However, some divergent cases occur where GPT-4o detects neuroticism with 62\% precision, significantly higher than in other models. Also, it is worth noticing that detecting High Neuroticism in discourse between the GPT-4o vs. Deepseek is more challenging than the other two combinations. 
        
        We used DeepSeek\footnote{\url{https://huggingface.co/deepseek-ai}} as a judge for pairwise conversation analysis. While LLaMA-3.3 and Qwen-2.5 required refinement, DeepSeek proved unreliable, with over 40\% invalid responses, leading to its exclusion from Table \ref{tab:comparison}.

       
        \textbf{Analysis Across Conversations}:  Compared to GPT-4o vs. GPT-4o-mini and GPT-4o vs Llama-3.3, the accuracy of High trait prediction in Neuroticism and Extraversion was significantly lower for GPT-4o vs. Deepseek conversation for both participants 1 and 2. This suggests that while exploration of low Neuroticism and Extraversion is comparable to the other two conversations, the complexity increases when these domains are High in the GPT-4o vs. Deepseek conversations. While observing individual participants across all the conversations, the results tend to be constant among the judges meaning if GPT rates high Agreeableness to participant 1 in one conversation, other judge models are likely to present similar results. 




    This addresses \textbf{RQ1 \& RQ3}. We observed a conditional capability of these agents as judges to accurately classify  traits from the discourses. This is true within various traits and also for the High and Low classification of the traits. Also, this finding provides an impression of inconsistency and bias towards certain OCEAN traits more than others. 
    \subsection{Inter Model Agreement}

Table \ref{tab:fleiss_kappa} presents the Fleiss' Kappa statistics, measuring inter-model agreement on personality trait judgments for Participants 1 and 2 across all dialogues.

In Discourse 1, Agreeableness showed moderate agreement ($\kappa$ > 0.5) for both participants. Openness agreement was substantial for Participant 1 but moderate for Participant 2. Conscientiousness and Neuroticism exhibited fair to moderate agreement.  Notably, Extraversion showed the lowest agreement, indicating poor reliability in its assessment.

Discourse 2 revealed minimal Agreeableness agreement for Participant 1 but substantially higher agreement for Participant 2, highlighting fluctuations in judging this trait. Openness maintained moderate to substantial agreement. Conscientiousness and Extraversion agreement increased compared to Discourse 1, though Extraversion remained low overall.  Neuroticism agreement showed a reversed trend, with lower agreement for Participant 1 and higher for Participant 2.

In Discourse 3, Agreeableness agreement remained moderate. Openness agreement decreased drastically. Conscientiousness, Extraversion, and Neuroticism agreement was stable between participants but only slight to fair.

These results address \textbf{RQ2}, demonstrating inconsistent inter-model agreement on personality traits.  Agreeableness and Openness agreement fluctuated across dialogues. The consistently low Extraversion agreement indicates significant challenges in its reliable assessment.  This variability underscores the non-uniformity of personality alignment in LLMs, highlighting difficulties in achieving stable and interpretable personality-driven interactions.

\subsection{Discourse Alignment with assigned personality traits}

Figure \ref{fig:three_figures} presents the accuracy of personality trait depiction for Participants 1 and 2, measured using LIWC-22.  GPT-4o-mini achieved the highest accuracy for Agreeableness across all dialogues.  However, GPT-4o's Agreeableness accuracy decreased substantially (from ~68\% and ~65\% to ~52\%) when conversing with Deepseek than GPT-4o-mini and Llama-3.3, suggesting a potential shift in personality expression depending on the interlocutor, similar to human behaviour \cite{atherton2022stability}.

Openness was the trait least accurately represented in all dialogues, with a maximum accuracy of ~51\%. This suggests that expressing Openness is particularly challenging for these LLMs.  Llama-3.3 exhibited the highest Conscientiousness, while GPT-4o showed the highest Extraversion. However, these differences were not statistically significant, and trait expression varied depending on the conversational partner.  GPT-4o's Neuroticism depiction was most accurate when interacting with Llama-3.3. This variability in traits and conversational settings directly addresses \textbf{RQ3}, confirming that all OCEAN traits are not equally prominent in generated conversations.

When comparing pairwise dialogues, GPT-4o vs. GPT-4o-mini and GPT-4o vs. Llama-3.3 showed similar performance.  However, GPT-4o vs. Deepseek dialogues exhibited significantly different results.  We observed that Deepseek struggled to consistently follow instructions from the prompts (even though the prompts were minimally adapted across models). Deepseek's generated text was also the most inconsistent in length compared to other models, which may have contributed to the observed differences.

\section{Conclusion}



This paper provides a comprehensive evaluation of trait adherence in LLM agents engaged in dyadic conversations.  Our findings highlight the significant challenges in achieving consistent and interpretable personality-aligned interactions. While LLMs can be guided to exhibit certain personality traits, their ability to maintain these traits across dynamic conversations varies considerably.  Future work should explore more sophisticated methods for instilling and evaluating personality, investigating the impact of dialogue context and developing metrics for assessing the nuances of personality expression in LLMs.  Exploring fine-tuning strategies or reinforcement learning approaches for improving consistency would also be valuable.

\section{Limitations}
One of the key challenges in this study is the absence of a standardized benchmarking system that all evaluations adhere to, making direct comparisons across different approaches more difficult. While strict rules were enforced to structure the discourse, models did not always fully comply, occasionally deviating from expected dialogue patterns. Additionally, there is a risk of bias, as language models may incorporate their own implicit judgments into discussions, potentially influencing personality assessments. Another important consideration is the length of dyadic conversations, there is no widely accepted standard for how long a dialogue should be to ensure a reliable evaluation. This uncertainty raises questions about whether longer or shorter exchanges might yield different insights, adding a layer of complexity to the interpretation of results.



\section{Ethical Considerations}

We do not collect any personal information and views for the creation of the discourse dataset or refer to any kind of personal traits from any sources to judge the nature of conversations. All the discourses are created by LLM agents. Topics provided for discussion for the agents are debatable but do not involve or promote the thought of violence, hatred or extremism of any kind to anyone. 

We use open and closed-source models that are available off the self and accessible to the general public. No changes in the model architecture have been made. Some hyperparameters have been adjusted to meet our expectations of the results, but they have been mentioned clearly in the paper. LLMs have the possibility of introducing bias in their results as per numerous studies. The dataset generated by the conversing agents has not been made public, but we do plan to publish it for further studies with careful ethical consideration and approvals. The results do present bias in predicting the BFI from the discourses but are solely limited to LLMs as judges. 

The content of LLM agents is subject to change if they are altered, fine-tuned, and tempered in different ways, which is a potential risk. 


\bibliography{custom}
\newpage
\appendix

\label{sec:appendix}

\section{Sample of Topics and Trait Combinations Used}
Samples of \emph{topics} used for debate:
\begin{lstlisting}
"Is the concept of a universal language beneficial?",
"Should the government regulate the pharmaceutical industry?",
"Is the use of nuclear energy justified?",
"Should the government provide free public transportation?",
"Is the concept of a cashless society beneficial?",
"Should the government regulate the gaming industry?"
\end{lstlisting}

\emph{Trait} combinations samples to assign personas to Agents: 

\begin{lstlisting}
{"Agreeableness": "High", "Openness": "Low", "Conscientiousness": "High", "Extraversion": "Low", "Neuroticism": "High"},
{"Agreeableness": "Low", "Openness": "High", "Conscientiousness": "Low", "Extraversion": "High", "Neuroticism": "Low"},
{"Agreeableness": "High", "Openness": "High", "Conscientiousness": "Low", "Extraversion": "High", "Neuroticism": "High"},
{"Agreeableness": "Low", "Openness": "Low", "Conscientiousness": "High", "Extraversion": "Low", "Neuroticism": "Low"},
{"Agreeableness": "High", "Openness": "High", "Conscientiousness": "High", "Extraversion": "Low", "Neuroticism": "Low"}
\end{lstlisting}

\section{System and User prompts}

We use, different \emph{System and User} prompts to extract the discourses and ratings from the conversing and judge agents. 

\subsection{Discourse Generation }
The \emph{system prompt} to generate the discourses:

\begin{lstlisting}
SYSTEM_PROMPT = ''' f"You are participating in a structured debate on: '{topic}'\n"
"Your responses should reflect these personality traits:\n"
f"- Agreeableness: {traits['Agreeableness']}\n"
f"- Openness: {traits['Openness']}\n"
f"- Conscientiousness: {traits['Conscientiousness']}\n"
f"- Extraversion: {traits['Extraversion']}\n"
f"- Neuroticism: {traits['Neuroticism']}\n\n"
"Rules:\n"
"- Maintain these personality traits (DO NOT EXPLICITLY MENTION IN TEXT) at all
times during your conversation\n"
"- Keep responses under 50 words\n"
"- Maintain your personality consistently\n"
"- Address previous arguments directly but do not repeat what
the other speaker said.\n"
"- End with proper punctuation" ''''
\end{lstlisting} 

The \emph{user prompt} carries the previous argument :
\begin{lstlisting}
USER_PROMPT = """Previous Argument:f"{previous_arguement}" """
\end{lstlisting}

\subsection{Extracting Personalities from the Judge Agents.}
The \emph{system prompt} to extract the personality traits:

\begin{lstlisting}
SYSTEM_PROMPT = """Analyze text segments from two anonymous debaters (Person One and Person Two) for:
1. Big Five Inventory (BFI) traits (High/Low for each dimension)
2. Consistency with typical behavior for those traits (Yes/No)

For each person, return:
{
    "predicted_bfi": {
        "Agreeableness": "High/Low",
        "Openness": "High/Low",
        "Conscientiousness": "High/Low",
        "Extraversion": "High/Low",
        "Neuroticism": "High/Low"
    }
}
"""
\end{lstlisting}
The \emph{user prompt} is: 

\begin{lstlisting}
USER_PROMPT= '''f"Analyze{persona}'s text:\n{text}'''    
\end{lstlisting}
where the \emph{persona} contains Participant 1 and 2  and the \emph{text} contains the discourses for each of the participants respectively. 

\newpage
\section{Metadata of the Discourses.}

\begin{table}[h]
    \centering
    \begin{tabular}{lc}
        \toprule
        \textbf{Metric} & \textbf{GPT-4o vs GPT-4o-mini} \\
        \midrule
        Total Sentences & 70,750 \\
        Total Words & 781,330 \\
        Assertions & 14,653 \\
        Questions & 1,507 \\
        Logical Structures & 690 \\
        Total Dialogues & 2,020 \\
        Avg. Words per Sentence & 11.04 \\
        Avg. Utterance Length & 48.35 \\
        \bottomrule
    \end{tabular}
    \caption{Metadata analysis for GPT-4o vs GPT-4o-mini}
    \label{tab:gpt4o_vs_gpt4o_mini}
\end{table}

\begin{table}[h]
    \centering
    \begin{tabular}{lc}
        \toprule
        \textbf{Metric} & \textbf{LLaMA-3 vs GPT-4o} \\
        \midrule
        Total Sentences & 44,964 \\
        Total Words & 541,603 \\
        Assertions & 15,577 \\
        Questions & 2,603 \\
        Logical Structures & 767 \\
        Total Dialogues & 2,020 \\
        Avg. Words per Sentence & 12.05 \\
        Avg. Utterance Length & 29.79 \\
        \bottomrule
    \end{tabular}
    \caption{Metadata analysis for LLaMA-3 vs GPT-40}
    \label{tab:llama3_vs_gpt4}
\end{table}

\begin{table}[h]
    \centering
    \begin{tabular}{lc}
        \toprule
        \textbf{Metric} & \textbf{DeepSeek vs GPT-4o} \\
        \midrule
        Total Sentences & 44,387 \\
        Total Words & 1,033,592 \\
        Assertions & 17,800 \\
        Questions & 380 \\
        Logical Structures & 4,697 \\
        Total Dialogues & 2,020 \\
        Avg. Words per Sentence & 23.29 \\
        Avg. Utterance Length & 56.85 \\
        \bottomrule
    \end{tabular}
    \caption{Metadata analysis for DeepSeek vs GPT-4o}
    \label{tab:deepseek_vs_llama}
\end{table}

\end{document}